\documentclass[conference]{IEEEtran}
\IEEEoverridecommandlockouts
\usepackage{cite}
\usepackage{amsmath,amssymb,amsfonts}
\usepackage{algorithmic}
\usepackage{graphicx}
\usepackage{textcomp}
\usepackage{balance}
\graphicspath{{images/}} 
\def\BibTeX{{\rm B\kern-.05em{\sc i\kern-.025em b}\kern-.08em
    T\kern-.1667em\lower.7ex\hbox{E}\kern-.125emX}}
\begin{document}


\title{Expectation Learning for Adaptive \\Crossmodal Stimuli Association\\
{\footnotesize }
\thanks{ 
}
}

\author{\IEEEauthorblockN{Pablo Barros$^1$,
German I. Parisi$^1$, Di Fu$^{2,3}$, Xun Liu$^{2,3}$, and Stefan Wermter$^1$}
\IEEEauthorblockA{$^1$Knowledge Technology, Department of Informatics, University of Hamburg, Hamburg, Germany}
\IEEEauthorblockA{$^2$CAS Key Laboratory of Behavioral Science, Institute of Psychology, Beijing, China}
\IEEEauthorblockA{$^3$Department of Psychology, University of Chinese Academy of Sciences, Beijing, China}}

\maketitle

\begin{abstract}

The human brain is able to learn, generalize, and predict crossmodal stimuli.
Learning by expectation fine-tunes crossmodal processing at different levels, thus enhancing our power of generalization and adaptation in highly dynamic environments.
In this paper, we propose a deep neural architecture trained by using expectation learning accounting for unsupervised
learning tasks.
Our learning model exhibits a self-adaptable behavior, setting the first steps towards the development of deep learning architectures for crossmodal stimuli association.

\end{abstract}

\section{Introduction}

Crossmodal processing is one of the characteristics of the human brain which is necessary for understanding the world around us.
The meaningful processing of crossmodal information allows us to enhance our perceptual experience \cite{Patton2003} also for unisensory stimuli \cite{Frassinetti2002}, to solve associative incongruence and conflicts \cite{Daconescu2011}, and to learn new concepts \cite{Dorst2001}.
 


Computational models for crossmodal learning have been proposed in the past to enhance tasks such as classification, regression, and prediction. Most of these models propose solutions for crossmodal fusion at an early \cite{Wei2010} or late stage \cite{Liu2016, deBoer2016}, e.g., by using crossmodal representations to increase the level of abstraction for a perception task. However, these models typically rely on individual and independent mechanisms for processing unimodal representations where modalities do not influence each other \cite{Poria2017, Chen2017}. Neurophysiological findings evidence that different brain regions are activated and communicating with each other also when processing unisensory information~\cite{Kayser2015}. The development of computational models that use brain-inspired principles for crossmodal processing may lead to more robust perception and interaction mechanisms in complex crossmodal environments.

In addition to the interaction between modality-specific regions, the brain also fine-tunes information using what is known as the expectation effect \cite{Yanagisawa2016}. While we are looking at an object, we are also estimating thousands of comparisons and clustering with similar and different objects that we have seen before. An even stronger effect of learning by expectation occurs with crossmodal information. When looking to a woman, one already expects to hear a female voice when she speaks~\cite{Tomlin2016}. This also causes an overfitting behavior when we have experienced very few examples in our lives: if we grew up near an opera house and never heard any other music style, every time we see a live show we would expect the singer to sing an opera. This is an important effect of learning as once we realize that there is an incongruence between what we expect and what really occurs, e.g. when the singer suddenly starts to sing death metal and not opera, we learn a novel association \cite{Ashby2016}. Such a learning process, referred to as \textit{learning by expectation}, makes us experts in associating concepts in an unsupervised way and using the difference of what was expected and what was perceived as a modulatory effect for learning new concepts.

In this paper, we introduce the use of a deep neural structure for crossmodal associative learning based on expectation. We evaluate our model on a crossmodal person identification task using visual and auditory information. The proposed model is composed of two channels, one to process each modality and composed of a series of convolution and pooling layers able to learn a low-dimensional representation of high-level abstract data. We use a self-organizing network to learn concurrent events for the two modalities. Additionally, each channel is paired with a reconstruction channel that can reconstruct the high-level input stimuli by using the low-dimensional specific representation. In contrast to conventional deep learning models, our model is trained entirely in an unsupervised fashion and has a continuous learning behavior. In other words, we do not rely on a large amount of data for learning and we also exhibit a self-adaptive behavior for the application of our learning model in real-world scenarios.



\section{Expectation Learning Model}

The proposed model is composed of two modality channels: one to learn facial features and the other to learn vocal characteristics. Each of these channels has two columns of convolution and pooling layers: one for perception and one for expectation. In the visual channel, the perception column is composed of three convolution layers, each of them followed by a pooling layer. At the end, we have a fully connected layer, representing the low-dimensional visual perception. The visual expectation column starts with a fully connected layer, with the same concept as in the perception column. We then proceed to have a similar structure as the visual perception column, but inverted. That means that instead of pooling, we use up-sampling operations to expand the dimensions of the information. In the end, we have an extra convolution layer with 3 filters to generate an RGB image.

 

Our auditory channel has a similar structure, with also two columns: perception and expectation. The perception column has three convolution layers. Each of these layers is followed by a max-pooling operator. This column has a fully connected layer that represents our low-dimensional auditory representation. We use 1 s of audio which is pre-processed using a 25 ms Hamming window with a stride of 20 ms and we produce a spectrogram using a log-Fourier transform with 40 Mel scale coefficients, producing a 100x40 spectrogram. The expectation column follows the same structure but inverted as in the visual column. We also make use of the subsampling instead of the max-pooling to expand the dimensionality of the data.



\begin{figure} 
	\center{\includegraphics[width=0.70\linewidth]{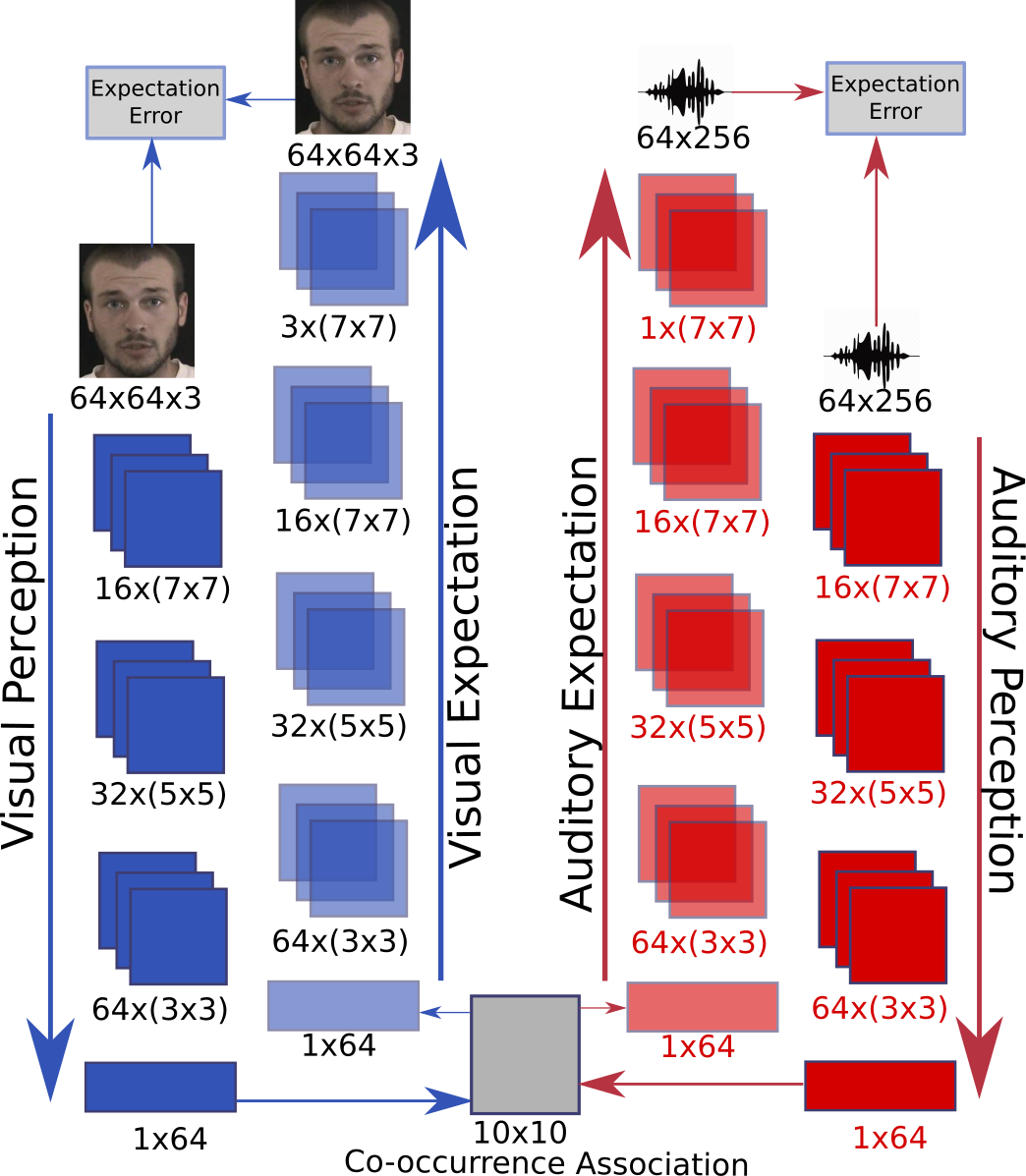}}
	\caption[FinalModel.]{Proposed model for crossmodal learning. The model has two modality-specific channels which have a perception and expectation columnn. We use a self-organizing co-occurrence association model to create prototype crossmodal neurons which will be used to generate expected stimuli.}
	\label{fig:finalModel}
\end{figure}

Each of the channels is connected to an unsupervised layer trained as a self-organizing map. This layer represents our co-occurrence association and is trained using both visual and auditory low-dimensional representation as input. We concatenate both representations and train the co-occurrence layer for each forward pass of the network. This allows us to create prototype neurons which will encode the crossmodal representation even when only one modality is present. All of our max-pooling layers have a dimension of 2x2. Figure \ref{fig:finalModel} shows the proposed model and architecture.

To train the co-occurrence layer, we first do a forward pass in the network to generate low-dimensional stimuli representations. Then, we train the co-occurrence association using the concatenated stimuli as input. For not biasing this layer towards representing the latest stimuli, we make use of a replay memory. This memory stores the 50 last forward passes. This helps to maintain the state of the self-organizing map based on previous stimuli and helps it to forget outliers. If an association is incongruent, but still present in the input data, it will be made fewer times than congruent associations and thus will be removed from the replay memory after a number of passes. We proceed to train this layer for 100 epochs for each forward pass using the whole replay memory as input.

Finally, to train the visual and auditory channels, we use an adversarial learning approach. Once we perform a forward pass, we use the best-matching unit of the co-occurrence layer to retrieve two low-dimensional representations: one visual and one auditory. We use this as input to both expectation models. Each expectation model will generate an expected stimulus. We then proceed to calculate the training loss which will be used to train our model, using the approach based on the Generative Adversarial Network (GAN) objective.

The Generative Adversarial Network (GAN) objective training can be explained as a generalization of our expectation learning. In GAN learning, the two entities, the discriminator and the generator are trained together to learn how to distinguish fake and original data and to generate indistinguishable fake data, respectively. One of the common problems of such models is the stabilization of the discriminator as the generator starts to learn how to copy the input data. The use of auto-encoders as discriminative/generative models was introduced and achieved better stabilization than traditional GANs \cite{Makhzani2015}, specifically because of the use of the auto-encoder loss to train the model instead of the categorical fake/not fake error. Our model uses the concept of a generative auto-encoder. The perception column can be explained as an encoding network and the expectation model as a decoding one.

Recent work on generative auto-encoders propose the use of the Wasserstein distance between the loss of the auto-encoder reconstruction for real and generated data as a loss function \cite{Berthelot2017}. Our approach uses a similar concept but instead of using the loss of an auto-encoder reconstruction, we use the Wasserstein distance between a perceived stimuli and an expected one. Due to the process of co-occurrence association, the expected stimuli can be reconstructed also when only one modality is present. This allows us to train both channels even when only one channel is perceived by using the unimodal expectation error.

\section{Preliminary Experiments and Results}
We evaluated our model using a crossmodal person identification task. We use the Enterface’05 audio-visual challenge database \cite{Martin2006} which contains recordings of 44 different subjects which are speaking phrases in different emotion intonations. Each subject has 5 examples of each of the 6 emotional intonations, for a total of 30 videos with a frame rate of 25 frames per second. We feed our network using always 1 s of audio and 1 frame which is randomly chosen from the 25 frames.

We trained our model with two different strategies: first, we trained the model with 70\% of the data of each subject and proceed to evaluate if the model recognizes the person using the other 30 \%  of the data. We proceeded with experiments with unimodal and multimodal stimuli. In the second strategy, we pre-trained the model with 20 subjects. We then proceeded to evaluate how the model learns the association of a new subject. This was done by analyzing the expectation error over time to see how it behaves while the network is learning a new association. For this purpose, we used a 6-second video with a subject which was not present in the training set.

For a proper evaluation, we used the person re-identification architecture by Ahmed et al. \cite{Ahmed2015}. This implementation uses multi-channel convolution layers to identify if two persons used as input are the same or not. We slightly modified the architecture using our auditory channel topology to make an auditory identification. As a baseline, training this network with the same data distribution as our first training strategy obtained an accuracy of 98\% for vision and 96\% for audio.
For this experiment, we calculated the reconstructed prediction based on crossmodal and unimodal stimuli.
The results of the first experiment are listed in Table \ref{tab:trainingStrategy1}.

\begin{table}
	\caption{Unimodal and Crossmodal accuracy and standard deviation for the first training strategy: training the crossmodal network for person identification.}
	\center \begin{tabular}{ c c c c c }
		\hline
		Auditory & Visual & Crossmodal \\ \hline
		91.4 \% (2.1) & 95.3 \% (2.2) & 98.2 \% (1.5)\\  
	\end{tabular} 
	\label{tab:trainingStrategy1}
\end{table}

\begin{figure} 
	\center{\includegraphics[width=0.8\linewidth]{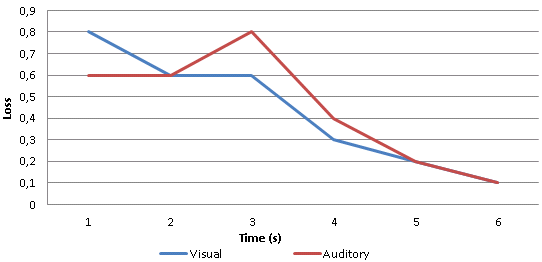}}
	\caption[FinalModel.]{Expectation error over time learning a new audio-visual stimulus for each channel.}
	\label{fig:plotLoss}
\end{figure}

The expectation error over time for a new subject is exhibited in Figure \ref{fig:plotLoss}. It is possible to see that the error for the visual expectation decreases faster than the auditory one although the auditory expectation error is lower as the first second is fed to the network. That means that the network associated better auditory characteristics than visual ones at first but later the visual channel learns faster. An interesting behavior is that the loss of the auditory stimulus grows after the third epoch, which could be explained by the update of the co-occurrence layer that is adapting to the new visual stimulus faster than the auditory one. 


\section{Conclusion and Future Work}

This paper shows the first steps towards the development of a deep neural network for learning crossmodal stimuli association via expectation learning.
Our network is trained fully unsupervised and adapts some characteristics of adversarial learning to simulate the expectation effect present in the human brain.
We stress the importance of having a self-adaptive learning model which does not rely on labeled training data.
Here, we showed preliminary experiments for a person identification task with unimodal and crossmodal stimuli and showed the behavior of the network when learning novel associations.

The proposed model does not take into consideration the crossmodal expectation error during training which can be modulated by unimodal stimuli.
We are planning to use unimodal stimuli as modulatory feedback for crossmodal learning which would improve the stability of the model and its power of generalization. We are also exploring the use of plasticity strategies for our unsupervised co-occurrence layer which can improve the learning capabilities of the network, currently limited by the topology of the self-organizing layer. 

\section*{Acknowledgements}
This work was partially supported by German Research Foundation (DFG) under project CML (TRR 169) and the NSFC (61621136008) and China Scholarship Council.

\bibliographystyle{IEEEtran}
\bibliography{bib}

\balance

\end{document}